\documentclass[10pt,twocolumn,letterpaper]{article}

\usepackage[pagenumbers]{wacv} % To force page numbers, e.g. for an arXiv version

\usepackage{algorithm}
\usepackage{algpseudocode}

\usepackage{float}
\usepackage{caption}
\usepackage{graphicx}
\usepackage{booktabs}
\usepackage{amsmath,amssymb} % define this before the line numbering.
\usepackage{color}
\usepackage{multirow}
\usepackage[normalem]{ulem}
\useunder{\uline}{\ul}{}
\usepackage{booktabs}
\definecolor{wacvblue}{rgb}{0.21,0.49,0.74}
\usepackage[pagebackref,breaklinks,colorlinks,allcolors=wacvblue]{hyperref}

\def\wacvPaperID{19} % *** Enter the WACV Paper ID here
\def\confName{WACV}
\def\confYear{2027}

\title{TTSA3R: Training-Free Temporal-Spatial Adaptive Persistent State for Streaming 3D Reconstruction}

\author{Zhijie Zheng, Xinhao Xiang, Jiawei Zhang\\
University of California, Davis\\
CA 95616, USA\\
{\tt\small \{zhjzheng, xhxiang, jiwzhang\}@ucdavis.edu}
}

\begin{document}

\twocolumn[{
  \maketitle
  \begin{center}
    \includegraphics[width=0.72\textwidth]{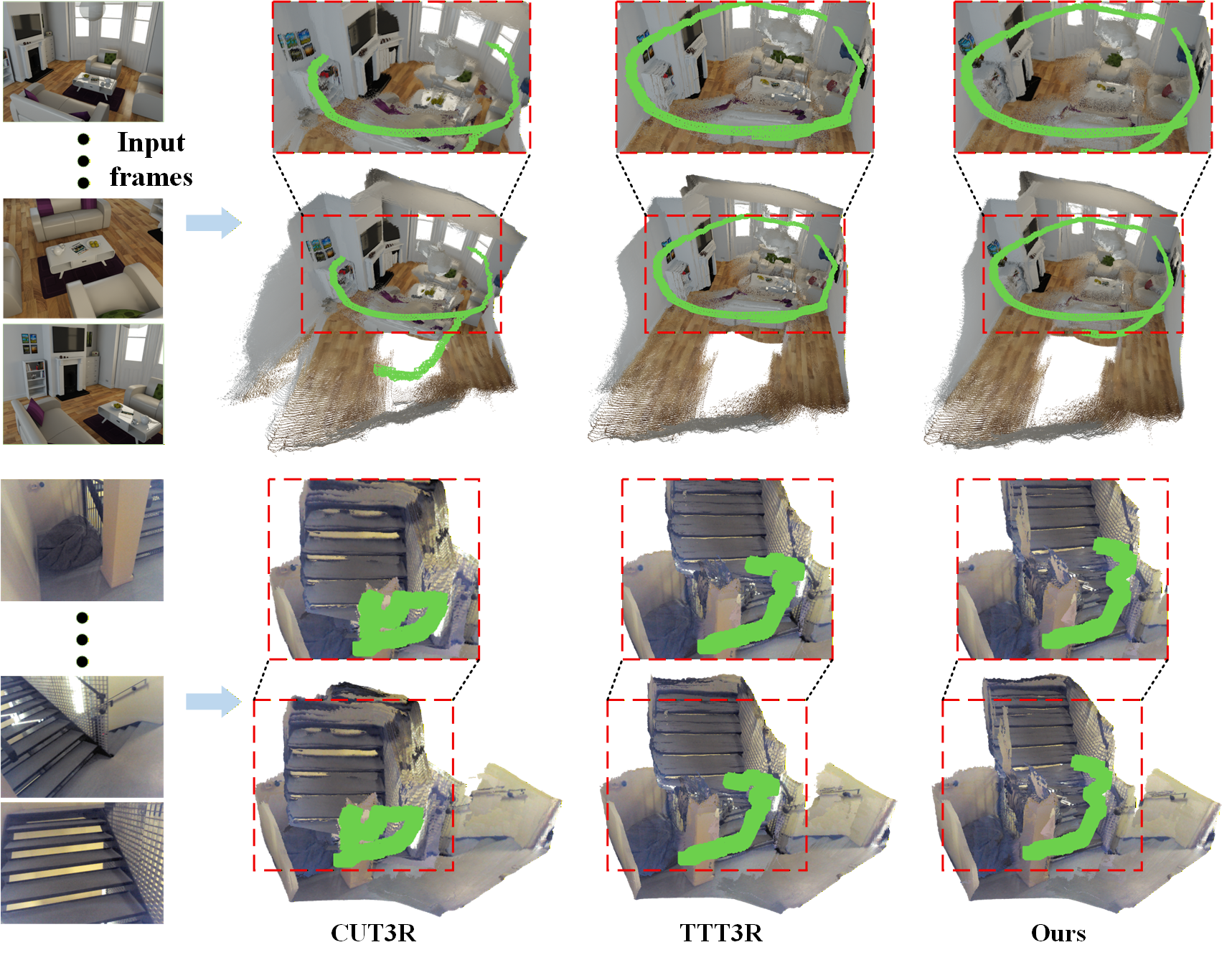}
    \captionof{figure}{\textbf{Catastrophic forgetting in streaming 3D reconstruction.} Given a sequence of input frames, CUT3R \cite{wang2025continuous} with uniform updates suffers from severe pose drift and geometric distortions over long sequences (\textbf{left}). TTT3R \cite{chen2025ttt3r} improves robustness but still exhibits artifacts (\textbf{middle}). Our method alleviates these issues through temporal-spatial adaptive updates to achieve coherent 3D reconstruction with accurate camera poses (\textbf{right}).}
  \label{fig:illustration}
  \end{center}
}]

\maketitle
\begin{abstract}
% Streaming recurrent models enable efficient 3D reconstruction by maintaining persistent state representations. However, they suffer from catastrophic forgetting over long sequences due to balancing historical information with new observations. Recent methods alleviate this by deriving adaptive signals from attention perspective, but they operate on single dimensions without considering temporal and spatial consistency. To this end, we propose a training-free framework termed \textbf{TTSA3R} that leverages both temporal state evolution and spatial observation quality for adaptive state updates in 3D reconstruction. In particular, we devise a Temporal Adaptive Update Module that regulates update magnitude by analyzing temporal state evolution patterns. Then, a Spatial Contextual Update Module is introduced to localize spatial regions that require updates through observation-state alignment and scene dynamics. These complementary signals are finally fused to determine the state updating strategies. Extensive experiments demonstrate the effectiveness of TTSA3R in diverse 3D tasks. Moreover, our method exhibits only 1.33$\times$ error increase compared to over 4$\times$ degradation in the baseline model on extended sequences of 3D reconstruction, significantly improving long-term reconstruction stability. The code will be made publicly available.

Streaming recurrent models enable efficient 3D reconstruction by maintaining persistent state representations. However, they suffer from catastrophic forgetting over long sequences due to balancing historical information with new observations. Recent methods alleviate this by deriving adaptive signals from the attention perspective, but they operate on single dimensions without considering temporal and spatial consistency. To this end, we propose a training-free framework termed TTSA3R that leverages both temporal state evolution and spatial observation quality for adaptive state updates in 3D reconstruction. In particular, we devise a Temporal Adaptive Update Module that regulates update magnitude by analyzing temporal state evolution patterns. Then, a Spatial Context Update Module is introduced to localize spatial regions that require updates through observation-state alignment and observation changes. These complementary signals are finally fused to determine the state updating strategies. Extensive experiments show that TTSA3R achieves competitive performance on standard short-sequence benchmarks and provides substantially stronger robustness on extended sequences. On NRGBD, as sequences extend from 50 to 250 frames, TTSA3R exhibits only a 1.33$\times$ error increase, compared with 3.8$\times$ degradation for CUT3R. This highlights the practical value of temporal-spatial adaptive updates for long-term reconstruction stability.
Our code is available at \url{https://github.com/anonus2357/ttsa3r}.

\end{abstract}
    
\section{Introduction}

With the increasing demand for real-time 3D perception, understanding 3D scene structure from videos or images has become essential for applications including robotic manipulation and augmented reality, where spatial awareness enables safe interaction with the physical world. Traditional Structure-from-Motion (SfM) \cite{to2008nonrigid, shah2014multi, li2021pixel} and Multi-View Stereo (MVS) \cite{ha2007surface, hepp2018plan3d,chen2019point} pipelines obtain high-precision 3D reconstruction through iterative optimization but are limited by complex workflows and high computational cost. Recent transformer-based methods like DUSt3R \cite{wang2024dust3r} and VGGT \cite{wang2025vggt} have demonstrated strong performance by joint reasoning across all frames with dense global attention \cite{vas2017atte}, producing impressive reconstruction results on standard benchmarks. Nevertheless, the quadratic complexity of dense attention presents a fundamental limitation, as memory requirements scale quadratically with the number of frames, making these approaches impractical for the scenarios with long frame sequence inputs.

Streaming methods \cite{chen2025long3r, chen2025human3r, yuan2025test3r, li2025wint3r, wu2025point3r} have emerged as an alternative by processing frames incrementally to update the learned states without revisiting previous observations, enabling online reconstruction with reduced computational overhead. Existing approaches include spatial memory-based methods that anchor observations at 3D positions (Spann3R \cite{wang20243d}, Point3R \cite{wu2025point3r}) and causal transformer architectures with key-value caching (STream3R \cite{lan2025stream3r}, StreamVGGT \cite{zhuo2025streaming}), which achieve efficient processing but suffer from unbounded memory growth as sequences extend. Alternatively, the recurrent model CUT3R \cite{wang2025continuous} maintains compact persistent states with constant memory footprint and exhibits competitive accuracy. 

However, extended sequences reveal limited length generalization due to catastrophic forgetting \cite{li2024incre,chen2025sab3r,chen2025ttt3r}. The uniform state update strategy \cite{wang2025continuous} fails to prevent low-quality observations from overwriting historical information, which leads to accumulated errors and severe geometric distortions, as illustrated in Figure \ref{fig:illustration}. In addition, spatial correspondence between states and observations is equally critical for determining update quality. To be specific, spatial interaction depends on both cross-attention alignment and feature consistency across frames. High cross-attention to regions with changing features indicates active geometric refinement, whereas cross-attention to regions with stable features suggests converged representations. Without jointly modeling these complementary signals, spatially irrelevant updates will corrupt stable geometry. Recent methods \cite{chen2025ttt3r,shen2025mut3r} alleviate this issue through attention-based adaptive update mechanisms. However, these approaches only rely on single attention signals to guide state updates, which limits fine-grained state update control.

In this work, we present a 3D reconstruction framework TTSA3R that mitigates the aforementioned issue through explicit temporal-spatial adaptive updates. We observe that state degradation stems from indiscriminate updates that fail to distinguish between stable geometry requiring preservation and outdated regions needing refinement. Therefore, effective updates should consider the analyses of both temporal and spatial signals. Specifically, temporal state evolution reveals the magnitude of change; spatial correspondence distinguishes whether high attention indicates active refinement with feature change or stable geometry with feature consistency. To this end, our method introduces dedicated modules to analyze these complementary dimensions, thus enabling fine-grained updates that preserve valuable information while integrating necessary observations.

To summarize, our main contributions are listed below:

$\bullet$ We propose a novel framework TTSA3R to alleviate long-term catastrophic forgetting for online streaming 3D reconstruction.

$\bullet$ We design a Temporal Adaptive Update Module (TAUM) based on state evolution analysis to track temporal state changes, which enables preservation of stable information and adapts to dynamic change.

$\bullet$ We introduce a Spatial Context Update Module (SCUM) combining cross-attention alignment and feature consistency to identify update-worthy regions, so as to suppress updates that are not supported by observation changes.

$\bullet$ Extensive experiments on video depth estimation, camera pose estimation, 3D reconstruction and diverse benchmarks demonstrate that our method achieves competitive short-sequence accuracy and substantially improved long-sequence robustness with real-time efficiency.

\section{Related Work}

\textbf{Classical and Neural Rendering 3D Reconstruction.} Structure-from-Motion \cite{snavely2006photo,agarwal2011bng,wu2013towards,ni2007out,sch2016struc,cui2017hsfm,pi2018learning,liu2024robust} and SLAM \cite{davison2003real,du2006simul,da2007monoslam,newcombe2011dtam,cadena2017past,huang2024photo} recover camera parameters and 3D structure through geometric optimization, while NeRF \cite{mildenhall2021nerf,chen2023single,guo2024sup,wu2024reconfusion} and 3DGS \cite{kerbl20233d,ch2024pixelsp,zhang2024gs,yuan2025robust,xia2025groom,tang2025dronesplat} achieve photorealistic synthesis through per-scene optimization. However, they cannot generalize to novel scenes without retraining, motivating learning-based methods.

\noindent\textbf{Feed-Forward 3D Reconstruction.} Recent works develop feed-forward networks \cite{wang2024dust3r, wang2025vggt, wang2025faster} for generalizable 3D reconstruction. DUSt3R \cite{wang2024dust3r} regresses pointmaps from image pairs without explicit matching, MASt3R \cite{leroy2024grounding} adds pixel-level correspondences, and Fast3R \cite{yang2025fast3r} achieves real-time performance. VGGT \cite{wang2025vggt} processes all frames jointly through global self-attention, and FastVGGT \cite{shen2025fastvggt} reduces its cost via token merging. However, global attention scales quadratically with the number of frames, making them difficult to scale to long sequences and motivating streaming pipelines.

\noindent\textbf{Streaming Online 3D Reconstruction.} Current streaming methods provide a real-time and memory-efficient scheme for 3D reconstruction. Spann3R~\cite{wang20243d} maintains an external spatial memory that stores previous 3D predictions. CUT3R~\cite{wang2025continuous} introduces a recurrent transformer with persistent state tokens that accumulate information from sequential observations to achieve constant memory usage. Point3R~\cite{wu2025point3r} proposes explicit spatial pointer memory where each pointer is assigned a 3D position and aggregates nearby scene information. STream3R~\cite{lan2025stream3r} reformulates reconstruction as a decoder-only transformer problem using causal attention. WinT3R~\cite{li2025wint3r} proposes a sliding window mechanism combined with a compact camera token pool. LONG3R~\cite{chen2025long3r} introduces memory gating with a 3D spatio-temporal memory. LongStream~\cite{cheng2026longstream} predicts keyframe-relative poses and periodically refreshes its cache, while LASER~\cite{ding2026laser} converts an offline model into a streaming one through layer-wise scale alignment. However, these methods either require dedicated training or rely on caches and windows of past frames. For recurrent models with constant memory, periodic state reset~\cite{chen2025ttt3r,cheng2026horizonstream} limits error accumulation at the cost of discarding historical information. To avoid costly training budget, recent research has shifted to training-free approaches with the aim of utilizing adaptive signals during inference to alleviate this problem. TTT3R~\cite{chen2025ttt3r} derives a per-token learning rate from the alignment confidence between the persistent state and the current observation (Sec.~3.3 of~\cite{chen2025ttt3r}), which is computed independently at each timestep. MUT3R~\cite{shen2025mut3r} aggregates multilayer self-attention to extract motion cues and suppresses dynamic regions through attention gating, and this signal is likewise confined to a single timestep. Ren et al.~\cite{ren2026rethinking} further rescale the gate of TTT3R with a frame-level scalar derived from inter-frame feature changes, which is shared by all state tokens within a frame. None of these methods explicitly models per-token candidate-state evolution across consecutive frames, limiting their ability to distinguish a token that needs continued refinement from one that has already stabilized. Therefore, we propose to integrate temporal dynamics and spatial context at the token level.

\section{Method}
\label{sec:method}
\begin{figure*}[h!]
	%\vspace{-0.3cm}
	\centering
	\includegraphics[width=1\linewidth]{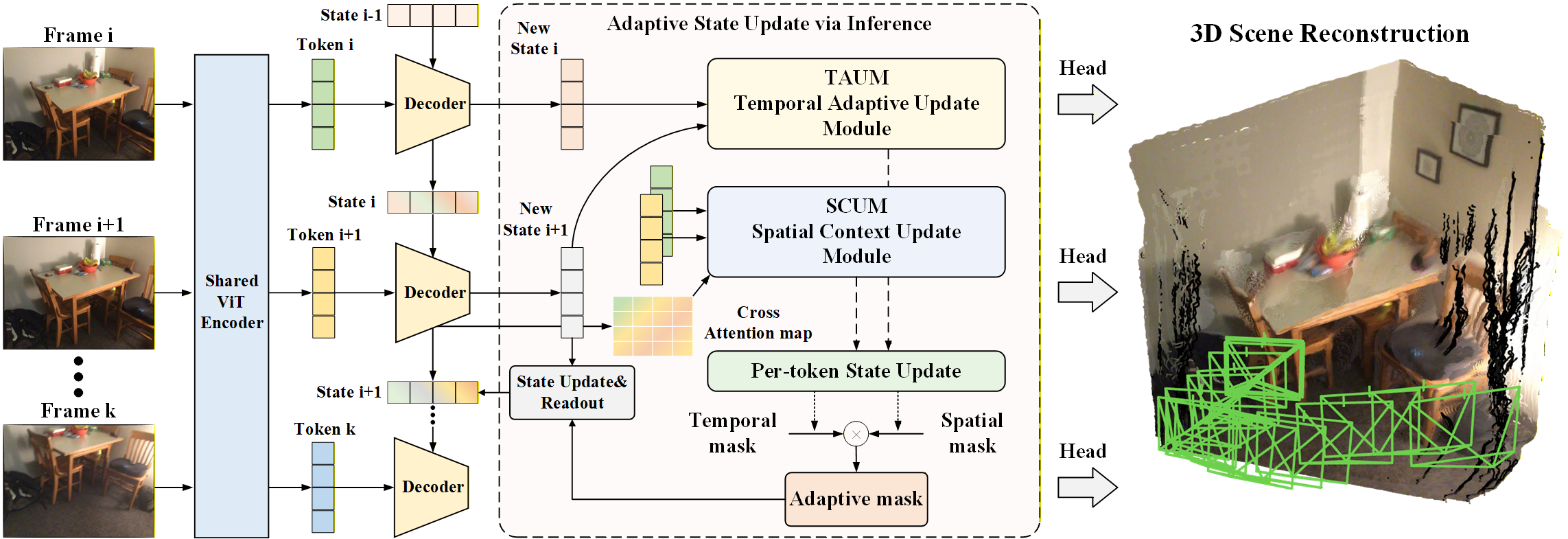}
	\caption{\textbf{Overview of our framework.} Our method performs streaming 3D reconstruction from sequential frames through adaptively updating persistent state. A shared ViT encoder extracts visual tokens from input frames, which interact with the persistent state via decoders to generate candidate states. Meanwhile, TAUM evaluates temporal state evolution across frames and SCUM measures spatial interaction between states and observations. Both modules produce complementary temporal and spatial masks, which are combined to update the persistent state at per-token granularity. Task-specific heads are utilized to predict depth maps, camera poses and pointmaps.}
	\label{fig:framework}
	\vspace{-0.3cm}
\end{figure*}

Our framework TTSA3R maintains a fixed-size persistent state that encodes scene geometry and produces per-frame 3D reconstructions. As illustrated in Figure \ref{fig:framework}, it consists of two main components: Temporal Adaptive Update Module and Spatial Context Update Module. These two modules provide complementary signals that are integrated for fine-grained state updates. %We first review CUT3R's inference in Section \ref{sec:pre}, and then detail TAUM in Section \ref{sec:taum} and SCUM in Section \ref{sec:scum}.

\subsection{Preliminaries: CUT3R Inference}
\label{sec:pre}
We first review the inference mechanism of CUT3R \cite{wang2025continuous}, which will be redesigned into an adaptive framework with the novel temporal and spatial update modules in this paper.
%which serves as the foundation for our adaptive framework. 
The existing CUT3R model adopts a recurrent architecture that has a compact global persistent state for streaming 3D reconstruction. Given an input image sequence $\{I_1, ..., I_T\}$, each frame $I_t$ is first encoded by models like ViT \cite{dos2020image} into image tokens:
\begin{equation}
    F_t = \text{Encoder}(I_t), \quad F_t \in \mathbb{R}^{K \times C}
\end{equation}
where $K$ denotes the number of patches and $C$ is the feature dimension.

Based on encoding results, a shared-weight transformer decoder \cite{wein2022croco, wang2024dust3r} is used to process the current image tokens $F_t$ together with the previous global state $S_{t-1} \in \mathbb{R}^{N \times C}$ and $N$ is the number of state tokens. The decoder outputs new state tokens $\tilde{S}_t$ and multi-layer features $G_t$:
\begin{equation}
    \tilde{S}_t, G_t = \text{Decoder}(F_t, S_{t-1})
\end{equation}
where $G_t = \left(G_t^{(0)}, ..., G_t^{(L)}\right )$ contains features from different decoder layers and $L$ is the total decoder depth. These multi-layer features will be fed into a prediction head to obtain geometric outputs:
\begin{equation}
    X_t, T_t, \text{conf}_t = \text{Head}(G_t)
\end{equation}
where $X_t$ denotes the predicted pointmap, $T_t$ is the camera pose and $\text{conf}_t$ represents per-pixel confidence.
The global persistent state is subsequently updated through masked interpolation:
\begin{equation}
    S_t = \tilde{S}_t \odot M + S_{t-1} \odot (1 - M)
\end{equation}
where $M \in \{0,1\}^{N \times 1}$ is a binary mask and $\odot$ denotes element-wise multiplication. In CUT3R, the mask is uniformly set to all ones, which directly replaces the previous state with new state tokens. Consequently, the global state is overwritten at every frame, and historical information is progressively lost as sequences become longer, leading to geometric drift.

\subsection{Temporal Adaptive Update Module}
\label{sec:taum}
To mitigate the catastrophic forgetting problem \cite{li2024incre,chen2025sab3r,chen2025ttt3r}, recent methods \cite{chen2025ttt3r,shen2025mut3r} capture adaptive updates from attention mechanisms. However, they mainly rely on intra-frame interactions within a single timestep, which lacks the ability to track inter-frame temporal changes. To this end, we propose the Temporal Adaptive Update Module (TAUM) to explicitly model how state representations vary and derive per-token update weights from temporal dynamics. 

The core principle of TAUM is to differentiate update strategies by measuring the temporal stability from state change magnitudes across consecutive frames, which selectively preserves stable information and adapts to changing observations. Specifically, tokens that exhibit minimal variation across frames have likely converged to reliable geometric representations and benefit from preserving historical information to maintain long-term consistency. In contrast, tokens with significant variation are still being refined by new observations and thus require larger updates. Since such variation may also arise from noisy estimates, TAUM is further combined with the spatial cues of SCUM in Sec.~\ref{sec:scum}.

In this paper, we formalize this principle through a per-token temporal adaptive mask as follows. To be specific, we measure the state evolution magnitude for each token and normalize the magnitude by the global average as:
\begin{equation}
    \begin{aligned}
    \Delta_t &= \text{Norm}(\tilde{S}_t - \tilde{S}_{t-1}) \\
    \hat{\Delta}_t &= \Delta_t \bigg/ \left( \frac{1}{N} \sum_{i=1}^N \Delta_t^{(i)} \right)
    \end{aligned}
\end{equation}
%where $\text{Norm}(\cdot)$ computes the element-wise $L2$ norm. 
where $\text{Norm}(\cdot)$ computes the per-token $L_2$-norm across feature dimensions.
Here we use candidate states $\tilde{S}_t$ rather than updated states $S_t$, since a token preserved in $S_t$ would show little change and remain frozen. For the first frame, where no previous frame exists, the state is directly set as $S_1=\tilde{S}_1$. Normalization further makes the change relative to the current scene. Then, we apply sigmoid gating to obtain the temporal mask as:
\begin{equation}
M_{\text{temp}} = \sigma (\hat{\Delta}_t - \tau)
\end{equation}
% where $\tau$ is a threshold parameter. 
where $M_{\text{temp}} \in \mathbb{R}^{N \times 1}$ and $\tau$ is a threshold parameter. We set $\tau=1.5$ for all experiments, with sensitivity analysis provided in Supp. Sec.~3. As $\hat{\Delta}_t$ has unit mean, this setting emphasizes tokens whose change is 50\% above the scene average.
Tokens with normalized changes above $\tau$ receive larger update weights to incorporate new information, while tokens below $\tau$ are down-weighted and largely retain their historical representations. Through this token-level adaptive mechanism, our design effectively alleviates the forgetting issue while maintaining the model's ability to capture dynamic scene changes.

\subsection{Spatial Context Update Module}
\label{sec:scum}

\begin{figure}[tbp]
	\centering
	\includegraphics[width=1.0\linewidth]{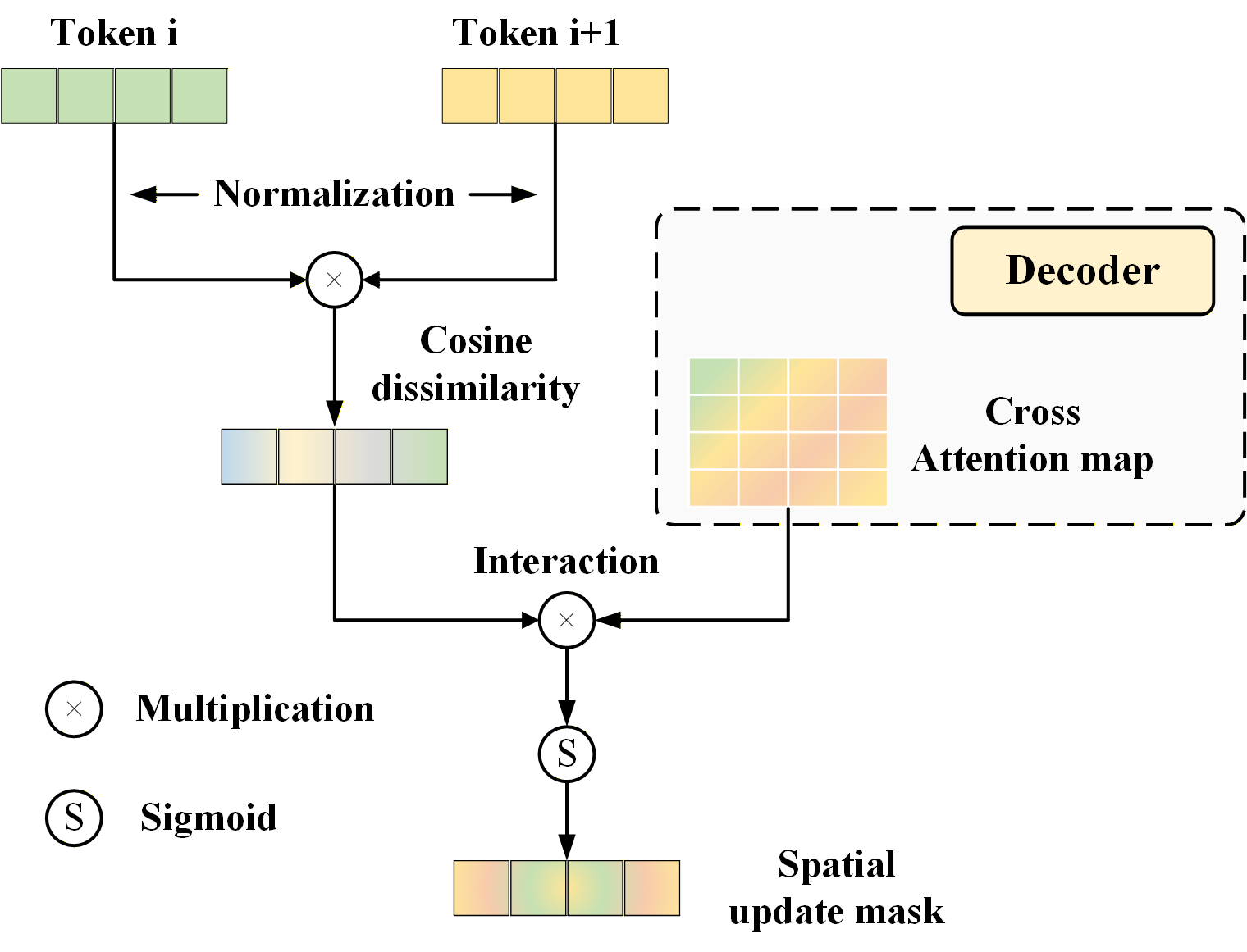}
	\caption{\textbf{Illustration of Spatial Context Update Module.} Spatial update masks are generated by combining cosine dissimilarity and cross-attention signals through multiplication and sigmoid activation.}
    \vspace{-0.3cm}
	\label{fig:SCUM}
\end{figure}

While TAUM mitigates temporal forgetting, it relies purely on temporal state tracking and neglects the spatial correspondence between states and observations. For instance, a token may change substantially across frames due to unreliable estimates, although the image regions it attends to remain unchanged. Therefore, we propose the Spatial Context Update Module (SCUM) to provide complementary spatial awareness, as illustrated in Figure \ref{fig:SCUM}.

Specifically, our key insight is that spatial interaction reveals update necessity through two complementary signals. Cross-attention captures alignment confidence \cite{chen2025ttt3r} between state tokens and image features. Additionally, feature divergence across consecutive frames reflects observation changes caused by either scene motion or viewpoint variation. Thus, high alignment confidence to regions with changing observations indicates active geometric refinement that requires updates. Otherwise, the update will be attenuated.

To achieve this, feature divergence is measured via cosine dissimilarity between consecutive frames as:
\begin{equation}
D_t = 1 - \text{CosSim}(F_t, F_{t-1})
\end{equation}
where the cosine similarity is computed along the feature dimension between image tokens at the same patch location, yielding $D_t \in \mathbb{R}^{1 \times K}$. In this way, $D_t$ captures new content caused by both scene and camera motion, which is further assigned to the attending state tokens via cross-attention.
% \textcolor{red}{State tokens $S_{t-1}$ serve as queries while image tokens $F_t$ serve as keys and values.}
In the decoder cross-attention, state tokens $S_{t-1}$ are used as queries, while image tokens $F_t$ are used as keys and values.
Cross-attention maps from the decoder layers are aggregated to quantify how strongly each state token engages with image observations as:
\begin{equation}
A_t = \Big| \frac{1}{LH} \sum_{l=1}^{L} \sum_{h=1}^{H} \text{Logit}^{(l,h)}({S}_{t-1}, F_t) \Big|
\end{equation}
where $\text{Logit}^{(l,h)} \in \mathbb{R}^{N \times K}$ denotes the pre-softmax cross-attention scores of head $h$ in decoder layer $l$, and $L$, $H$ are the numbers of layers and heads. The spatial interaction is computed as the element-wise product of attention and dissimilarity, where $D_t$ is shared across all state tokens, followed by max pooling across image tokens and sigmoid gating to derive the spatial mask as:
\begin{equation}
M_{\text{spat}} = \sigma(max_{\text{spat}}(A_t \odot D_t))
\end{equation}

The element-wise product $\odot$ ensures high activation only when both attention and divergence are substantial. Spatial max pooling $max_{\text{spat}}$ exploits the observation that each state token typically corresponds to specific image regions. Since $A_t$ and $D_t$ are non-negative, $M_{\text{spat}} \in \mathbb{R}^{N \times 1}$ lies in $[0.5, 1)$. Therefore, SCUM softly modulates the temporal update, retaining it for tokens engaging with changing observations and reducing it by up to half for tokens attending to unchanged regions.

Finally, in order to integrate these complementary perspectives, we combine TAUM and SCUM to obtain the adaptive mask as:
\begin{equation}
M_{\text{final}} = M_{\text{temp}} \odot M_{\text{spat}}
\end{equation}
% This fusion ensures updates only when both temporal dynamics and spatial correspondence are met. 
In this fusion, $M_{\text{temp}}$ determines the primary update weight and $M_{\text{spat}}$ further modulates it, so that the largest updates occur only when both temporal dynamics and spatial correspondence are present. We compare different fusion strategies in Supp. Sec.~2.
Then, the global state update is illustrated as:
\begin{equation}
S_t = \tilde{S}_t \odot M_{\text{final}} + S_{t-1} \odot (1 - M_{\text{final}})
\end{equation}
Since both signals are extracted from internal encoder and decoder representations, TTSA3R requires no additional training or forward passes. This enables the model to maintain long-term consistency across extended sequences.
\section{Experiments}
\label{sec:exp}

\textbf{Implementation Details.} %Our framework TTSA3R builds on the CUT3R architecture with pretrained weights. 
We conduct experiments on three tasks including video depth estimation, camera pose estimation, and 3D reconstruction. Following the common practice \cite{wang2025continuous,chen2025ttt3r,zhuo2025streaming}, we assess video depth on Sintel \cite{butler2012natc}, Bonn \cite{palao2019refun}, and KITTI \cite{geiger2013vision} datasets, where we report absolute relative error (Abs Rel) and the percentage of pixels with relative error below 1.25 ($\delta<1.25$). We then evaluate camera pose estimation on Sintel, TUM-dynamics \cite{sturm2012benchmark}, and ScanNet \cite{dai2017scannet} datasets, measuring Absolute Translation Error (ATE), Relative Translation Error (RPE$_{\text{trans}}$), and Relative Rotation Error (RPE$_{\text{rot}}$) after Sim(3) alignment with ground truth. For 3D reconstruction, we use NRGBD \cite{azinovic2022neural} dataset and report accuracy (Acc) as well as normal consistency (NC). We compare against methods grouped by different processing paradigms, including optimization-based methods DUSt3R-GA \cite{wang2024dust3r}, MASt3R-GA \cite{leroy2024grounding} and MonST3R-GA \cite{zhang2024monst3r}, full-attention methods Easi3R \cite{chen2025easi3r} and VGGT \cite{wang2025vggt}, as well as streaming methods Spann3R \cite{wang20243d}, Point3R \cite{wu2025point3r}, CUT3R \cite{wang2025continuous}, TTT3R \cite{chen2025ttt3r}, STream3R \cite{lan2025stream3r}, StreamVGGT \cite{zhuo2025streaming} and MUT3R \cite{shen2025mut3r}. For long sequences, we follow the evaluation protocol of TTT3R \cite{chen2025ttt3r}. All experiments are performed on a single NVIDIA A6000 GPU, where TTSA3R runs at 18.0 FPS with 6 GB memory, comparable to TTT3R (Supp. Sec.~5).

\begin{table*}[t]
\caption{\textbf{Video Depth Estimation} on standard short sequences. We evaluate scale-invariant and metric depth accuracy on Sintel \cite{butler2012natc}, Bonn \cite{palao2019refun}, and KITTI \cite{geiger2013vision} datasets. Methods that require global alignment are denoted as "GA". "Optim", "Stream", and "FA" refer to optimization-based, streaming, and full-attention methods, respectively. The best and second best results are marked in bold and underline.}
\label{table:1}
\resizebox{\linewidth}{!}{
\begin{tabular}{ccccc|cc|cc}
\toprule[1pt] %\hline
\multirow{2}{*}{\textbf{Alignment}}             & \multirow{2}{*}{\textbf{Method}} & \multirow{2}{*}{\textbf{Type}} & \multicolumn{2}{c}{\textbf{Sintel}}                              & \multicolumn{2}{c}{\textbf{Bonn}}                                & \multicolumn{2}{c}{\textbf{KITTI}}          \\  \cmidrule(lr){4-5} \cmidrule(lr){6-7} \cmidrule(lr){8-9} %\cmidrule[0.6pt]{4-9}%\cline{4-9} 
&      &       & Abs Rel $\downarrow$       & $\delta<1.25 \uparrow$  & Abs Rel $\downarrow$    & $\delta <1.25 \uparrow$ & Abs Rel $\downarrow$    & $\delta<1.25 \uparrow$ \\ \midrule[0.6pt] %\hline
\multirow{13}{*}{\textbf{Per-sequence   scale}} & DUSt3R-GA \cite{wang2024dust3r}               & Optim                 & 0.656          & \multicolumn{1}{c|}{45.2}              & 0.155          & \multicolumn{1}{c|}{83.3}              & 0.144          & 81.3              \\
                                       & MASt3R-GA \cite{leroy2024grounding}               & Optim                 & 0.641          & \multicolumn{1}{c|}{43.9}              & 0.252          & \multicolumn{1}{c|}{70.1}              & 0.183          & 74.5              \\
                                       & MonST3R-GA \cite{zhang2024monst3r}             & Optim                 & 0.378          & \multicolumn{1}{c|}{55.8}              & 0.067          & \multicolumn{1}{c|}{96.3}              & 0.168          & 74.4              \\
                                       & Easi3R \cite{chen2025easi3r}                 & FA                    & {\ul 0.377}    & \multicolumn{1}{c|}{{\ul 55.9}}        & {\ul 0.059}    & \multicolumn{1}{c|}{{\ul 97.0}}        & {\ul 0.102}    & {\ul 91.2}        \\
                                       & VGGT \cite{wang2025vggt}                   & FA                    & \textbf{0.287} & \multicolumn{1}{c|}{\textbf{66.1}}     & \textbf{0.055} & \multicolumn{1}{c|}{\textbf{97.1}}     & \textbf{0.070} & \textbf{96.5}     \\ \cmidrule[0.6pt]{2-9} %\cline{2-9} 
                                       & Spann3R \cite{wang20243d}                & Stream                & 0.622          & \multicolumn{1}{c|}{42.6}              & 0.144          & \multicolumn{1}{c|}{81.3}              & 0.198          & 73.7              \\
                                       & Point3R \cite{wu2025point3r}                & Stream                & 0.452          & \multicolumn{1}{c|}{48.9}              & {\ul 0.060}    & \multicolumn{1}{c|}{96.0}              & 0.136          & 84.2              \\
                                       & CUT3R \cite{wang2025continuous}                  & Stream                & 0.421          & \multicolumn{1}{c|}{47.9}              & 0.078          & \multicolumn{1}{c|}{93.7}              & 0.118          & 88.1              \\
                                       & TTT3R \cite{chen2025ttt3r}                  & Stream                & 0.405          & \multicolumn{1}{c|}{48.9}              & 0.069          & \multicolumn{1}{c|}{95.4}              & {\ul 0.114}    & {\ul 90.4}        \\
                                       & STream3R$^{\alpha}$ \cite{lan2025stream3r}                & Stream                & 0.478          & \multicolumn{1}{c|}{{\ul 51.1}}        & 0.075          & \multicolumn{1}{c|}{94.1}              & 0.116          & 89.6              \\
                                       & StreamVGGT \cite{zhuo2025streaming}             & Stream                & \textbf{0.323} & \multicolumn{1}{c|}{\textbf{65.7}}     & \textbf{0.059} & \multicolumn{1}{c|}{\textbf{97.2}}     & 0.173          & 72.1              \\
                                       & MUT3R \cite{shen2025mut3r}                  & Stream                & 0.451          & \multicolumn{1}{c|}{48.6}              & 0.070          & \multicolumn{1}{c|}{96.2}              & 0.116          & 88.3              \\
                                       & \textbf{Ours}           & Stream                & {\ul 0.401}    & \multicolumn{1}{c|}{50.0}              & 0.064          & \multicolumn{1}{c|}{{\ul 96.5}}        & \textbf{0.110} & \textbf{91.2}     \\ \midrule[0.6pt] %\hline
\multirow{7}{*}{\textbf{Metric scale}}          & MASt3R-GA \cite{leroy2024grounding}              & Optim                 & 1.022          & \multicolumn{1}{c|}{14.3}              & 0.272          & \multicolumn{1}{c|}{70.6}              & 0.467          & 15.2              \\
                                       & CUT3R \cite{wang2025continuous}                  & Stream                & 1.029          & \multicolumn{1}{c|}{23.8}              & 0.103          & \multicolumn{1}{c|}{88.5}              & 0.122          & 85.5              \\
                                       & Point3R \cite{wu2025point3r}                & Stream                & \textbf{0.777} & \multicolumn{1}{c|}{17.1}              & 0.137          & \multicolumn{1}{c|}{94.7}              & 0.191          & 73.8              \\
                                       & TTT3R \cite{chen2025ttt3r}                  & Stream                & 0.977          & \multicolumn{1}{c|}{24.5}              & 0.090          & \multicolumn{1}{c|}{94.2}              & \textbf{0.110} & \textbf{89.1}     \\
                                       & STream3R$^{\alpha}$ \cite{lan2025stream3r}                & Stream                & 1.041          & \multicolumn{1}{c|}{21.0}              & {\ul 0.084}    & \multicolumn{1}{c|}{94.4}              & 0.234          & 57.6              \\
                                       & MUT3R \cite{shen2025mut3r}                  & Stream                & {\ul 0.820}    & \multicolumn{1}{c|}{\textbf{25.2}}     & 0.086          & \multicolumn{1}{c|}{{\ul 96.0}}        & 0.125          & 85.8              \\
                                       & \textbf{Ours}           & Stream                & 0.960          & \multicolumn{1}{c|}{{\ul 24.6}}        & \textbf{0.079} & \multicolumn{1}{c|}{\textbf{96.6}}     & {\ul 0.111}    & {\ul 88.9}        \\ \bottomrule[1pt] %\hline
\end{tabular}
}
% \vspace{-0.3cm}
\end{table*}

\subsection{Video Depth Estimation}

We evaluate video depth on both short and long sequences. Table \ref{table:1} shows the results on standard benchmarks under two evaluation protocols, which are per-sequence scale and metric scale directly comparing absolute depth values. Under per-sequence scale, our method TTSA3R achieves the best performance among streaming methods on KITTI \cite{geiger2013vision} dataset while remaining competitive on the other two datasets, where StreamVGGT benefits from caching all past frames. Moreover, TTSA3R demonstrates superior performance on Bonn \cite{palao2019refun} dataset under metric scale.

\begin{figure}[t!]
	\centering
    %\vspace{-0.3cm}
	\includegraphics[width=1.0\linewidth]{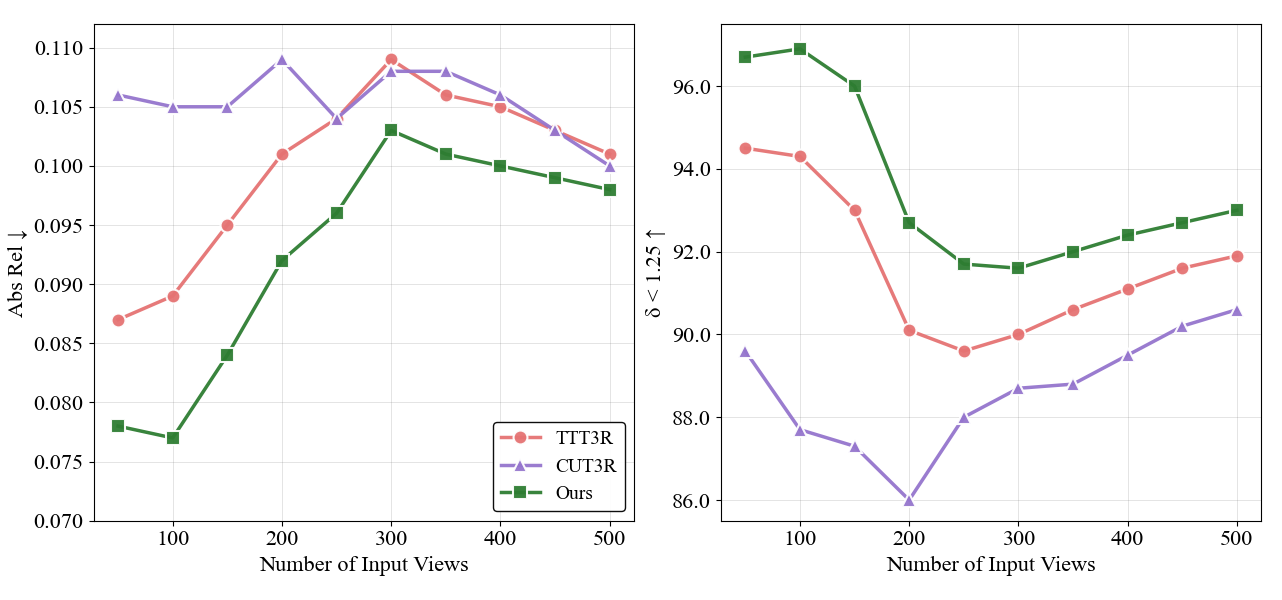}
	\caption{\textbf{Video depth estimation} (long sequences) using metric depth accuracy on Bonn \cite{palao2019refun} dataset.}
    \vspace{-0.3cm}
	\label{fig:BONN-long}
\end{figure}

Figure \ref{fig:BONN-long} illustrates how video depth evolves as input sequence length increases from 50 to 500 frames. It can be seen that TTSA3R achieves the lowest Abs Rel and the highest $\delta<1.25$ at every sequence length. TTT3R \cite{chen2025ttt3r} improves over CUT3R \cite{wang2025continuous} on shorter sequences, but its advantage in Abs Rel becomes marginal from 250 frames onward. On the contrary, TTSA3R retains a clear margin in $\delta<1.25$ over both methods throughout the sequences (93.0 vs.\ 91.9 and 90.6 at 500 frames). This illustrates that our temporal-spatial fusion preserves geometric fidelity over long sequences.

\begin{table*}[htbp!]
\caption{\textbf{Camera Pose Estimation} on standard short sequences. We evaluate three metrics on Sintel \cite{butler2012natc}, TUM-dynamics \cite{sturm2012benchmark}, and ScanNet \cite{dai2017scannet} datasets.}
\label{table:2}
\resizebox{\linewidth}{!}{
\begin{tabular}{ccccc|ccc|ccc}
\toprule[1pt]  %\hline
\multirow{2}{*}{\textbf{Method}} & \multirow{2}{*}{\textbf{Type}} & \multicolumn{3}{c}{\textbf{Sintel}}                                            & \multicolumn{3}{c}{\textbf{TUM-dynamics}}       & \multicolumn{3}{c}{\textbf{ScanNet}}        \\ \cmidrule(lr){3-5} \cmidrule(lr){6-8} \cmidrule(lr){9-11} %\cline{3-11}       
&          & ATE $\downarrow$        & RPE trans $\downarrow$     & RPE rot $\downarrow$     & ATE $\downarrow$           & RPE trans $\downarrow$     & RPE rot $\downarrow$       & ATE $\downarrow$           & RPE trans $\downarrow$     & RPE rot $\downarrow$       \\  \midrule[0.6pt] %\hline 
Robust-CVD \cite{kopf2021robust}             & Optim                 & 0.360          & 0.154          & \multicolumn{1}{c|}{3.443}          & 0.153          & 0.026          & \multicolumn{1}{c|}{3.528}          & 0.227          & 0.064          & 7.374          \\
CasualSAM \cite{zh2022structure}              & Optim                 & 0.141          & \textbf{0.035} & \multicolumn{1}{c|}{{\ul 0.615}}          & 0.071          & \textbf{0.010} & \multicolumn{1}{c|}{1.712}          & 0.158          & 0.034          & 1.618          \\
DUSt3R-GA \cite{wang2024dust3r}              & Optim                 & 0.417          & 0.250          & \multicolumn{1}{c|}{5.796}          & 0.083          & 0.017          & \multicolumn{1}{c|}{3.567}          & 0.081          & 0.028          & 0.784          \\
MASt3R-GA \cite{leroy2024grounding}              & Optim                 & 0.185          & 0.060          & \multicolumn{1}{c|}{1.496}          & {\ul 0.038}    & {\ul 0.012}    & \multicolumn{1}{c|}{{\ul 0.448}}    & 0.078          & 0.020          & {\ul 0.475}    \\
MonST3R-GA \cite{zhang2024monst3r}             & Optim                 & {\ul 0.111}    & 0.044          & \multicolumn{1}{c|}{0.869}          & 0.098          & 0.019          & \multicolumn{1}{c|}{0.935}          & 0.077          & 0.018          & 0.529          \\
Easi3R \cite{chen2025easi3r}                 & FA                    & \textbf{0.110} & {\ul 0.042}    & \multicolumn{1}{c|}{0.758}          & 0.105          & 0.022          & \multicolumn{1}{c|}{1.064}          & {\ul 0.061}    & {\ul 0.017}    & 0.525          \\
VGGT \cite{wang2025vggt}                   & FA                    & 0.172          & 0.062          & \multicolumn{1}{c|}{\textbf{0.471}} & \textbf{0.012} & \textbf{0.010} & \multicolumn{1}{c|}{\textbf{0.310}} & \textbf{0.035} & \textbf{0.015} & \textbf{0.377} \\ \midrule[0.6pt] %\hline
Spann3R \cite{wang20243d}                & Stream                & 0.329          & 0.110          & \multicolumn{1}{c|}{4.471}          & 0.056          & 0.021          & \multicolumn{1}{c|}{0.591}          & 0.096          & 0.023          & 0.661          \\
CUT3R \cite{wang2025continuous}                  & Stream                & {\ul 0.213}    & {\ul 0.066}    & \multicolumn{1}{c|}{\textbf{0.621}} & 0.046          & 0.015          & \multicolumn{1}{c|}{0.473}          & 0.099          & 0.022          & {\ul 0.600}          \\
Point3R \cite{wu2025point3r}                & Stream                & 0.351          & 0.128          & \multicolumn{1}{c|}{1.822}          & 0.075          & 0.029          & \multicolumn{1}{c|}{0.642}          & 0.106          & 0.035          & 1.946          \\
TTT3R \cite{chen2025ttt3r}                  & Stream                & \textbf{0.210} & 0.090          & \multicolumn{1}{c|}{{\ul 0.722}}    & {\ul 0.028}    & {\ul 0.013}    & \multicolumn{1}{c|}{{\ul 0.380}}    & {\ul 0.064}          & {\ul 0.021}    & 0.637          \\
MUT3R \cite{shen2025mut3r}                  & Stream                & 0.228          & \textbf{0.062} & \multicolumn{1}{c|}{0.751}          & 0.042          & 0.015          & \multicolumn{1}{c|}{0.445}          & -    & -          & - \\
\textbf{Ours}                    & Stream                & \textbf{0.210} & 0.085          & \multicolumn{1}{c|}{0.765}          & \textbf{0.026} & \textbf{0.012} & \multicolumn{1}{c|}{\textbf{0.372}} & \textbf{0.057} & \textbf{0.020} & \textbf{0.588}    \\ \bottomrule[1pt] %\hline
\end{tabular}
}
% \vspace{-0.3cm}
\end{table*}

\subsection{Camera Pose Estimation}

\begin{figure}[h!]
	\centering
    %\vspace{-0.3cm}
	\includegraphics[width=1.0\linewidth]{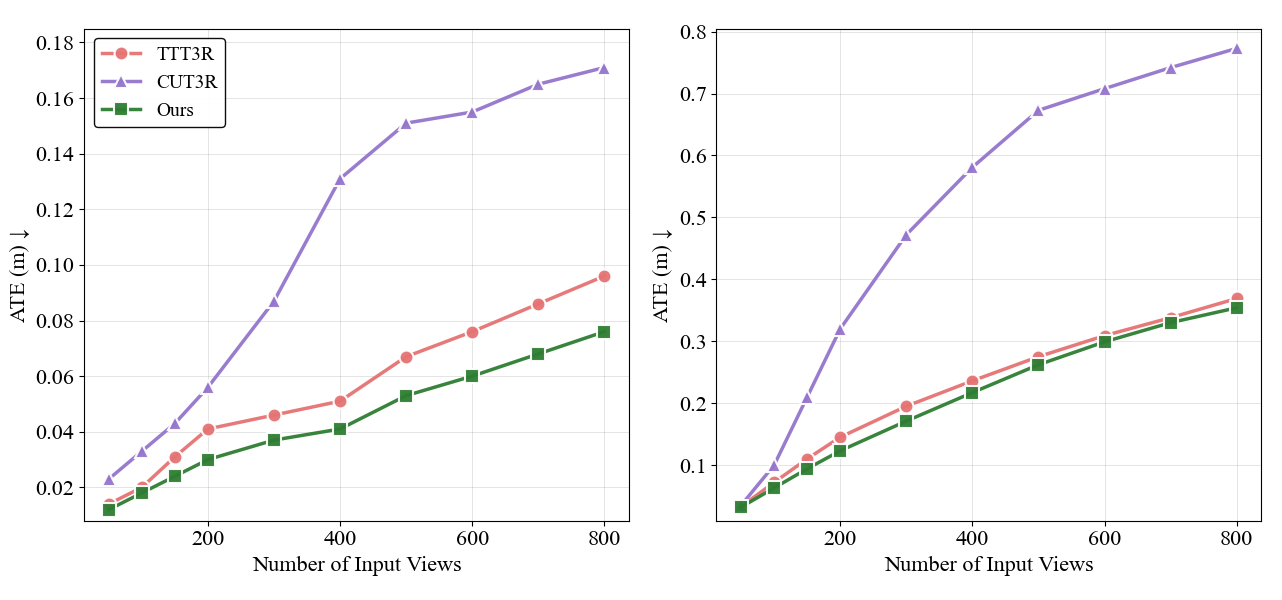}
	\caption{\textbf{Camera pose estimation} (long sequences) on TUM-dynamics \cite{sturm2012benchmark} (left) and ScanNet \cite{dai2017scannet} (right) datasets.}
    % \vspace{-0.3cm}
	\label{fig:cam-long}
\end{figure}

To validate the effectiveness of our method, we next conduct experiments on camera pose estimation. Table \ref{table:2} presents the quantitative results across Sintel \cite{butler2012natc}, TUM-dynamics \cite{sturm2012benchmark}, and ScanNet \cite{dai2017scannet} datasets. It is shown that TTSA3R achieves the lowest ATE among streaming methods on TUM-dynamics and ScanNet datasets.
It outperforms optimization-based pipelines like DUSt3R-GA \cite{wang2024dust3r} with global alignment and surpasses the full-attention baseline Easi3R \cite{chen2025easi3r} in ATE on TUM-dynamics and ScanNet. Besides, we also evaluate pose tracking on sequences up to 800 frames for long-term stability assessment, as illustrated in Figure \ref{fig:cam-long}. Across both datasets, our method TTSA3R achieves the lowest pose error at all sequence lengths, and its margin over CUT3R widens as sequences grow. 

\subsection{3D Reconstruction}

\begin{figure}[htbp]
	\centering
	\includegraphics[width=1.0\linewidth]{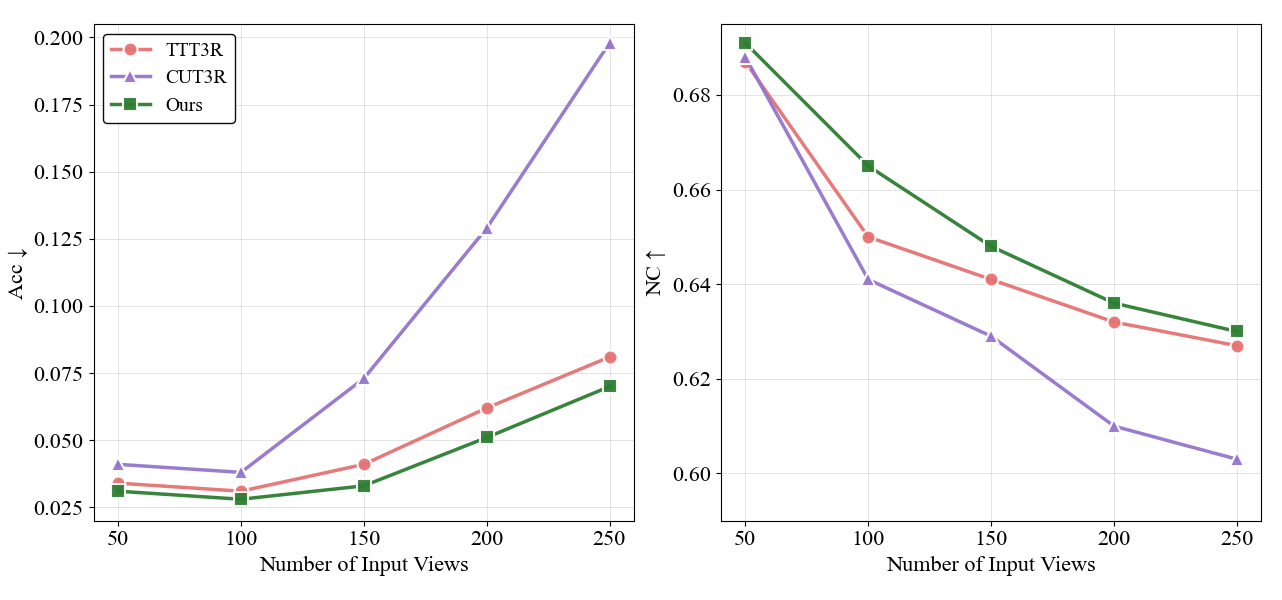}
	\caption{\textbf{3D reconstruction} (long sequences) on NRGBD \cite{azinovic2022neural} dataset.}
    %\vspace{-0.3cm}
	\label{fig:3DR-long}
\end{figure}

% We further evaluate 3D reconstruction on NRGBD \cite{azinovic2022neural} dataset across different sequences. The quantitative results are shown in Figure \ref{fig:3DR-long}. Compared to CUT3R \cite{wang2025continuous} and TTT3R \cite{chen2025ttt3r}, our method shows obviously better performance on all sequence lengths, which exhibits effective resistance to the catastrophic forgetting problem. 
We further evaluate 3D reconstruction on NRGBD \cite{azinovic2022neural} dataset, as shown in Figure \ref{fig:3DR-long}. Our method exhibits only 1.33x error increase compared to 3.8$\times$ degradation in baseline CUT3R \cite{wang2025continuous} as sequences extend from 50 to 250 frames. Compared to TTT3R \cite{chen2025ttt3r}, our method shows better performance on all sequence lengths, demonstrating more effective resistance to catastrophic forgetting.

\begin{figure*}[htbp]
	\centering
	\includegraphics[width=0.65\linewidth]{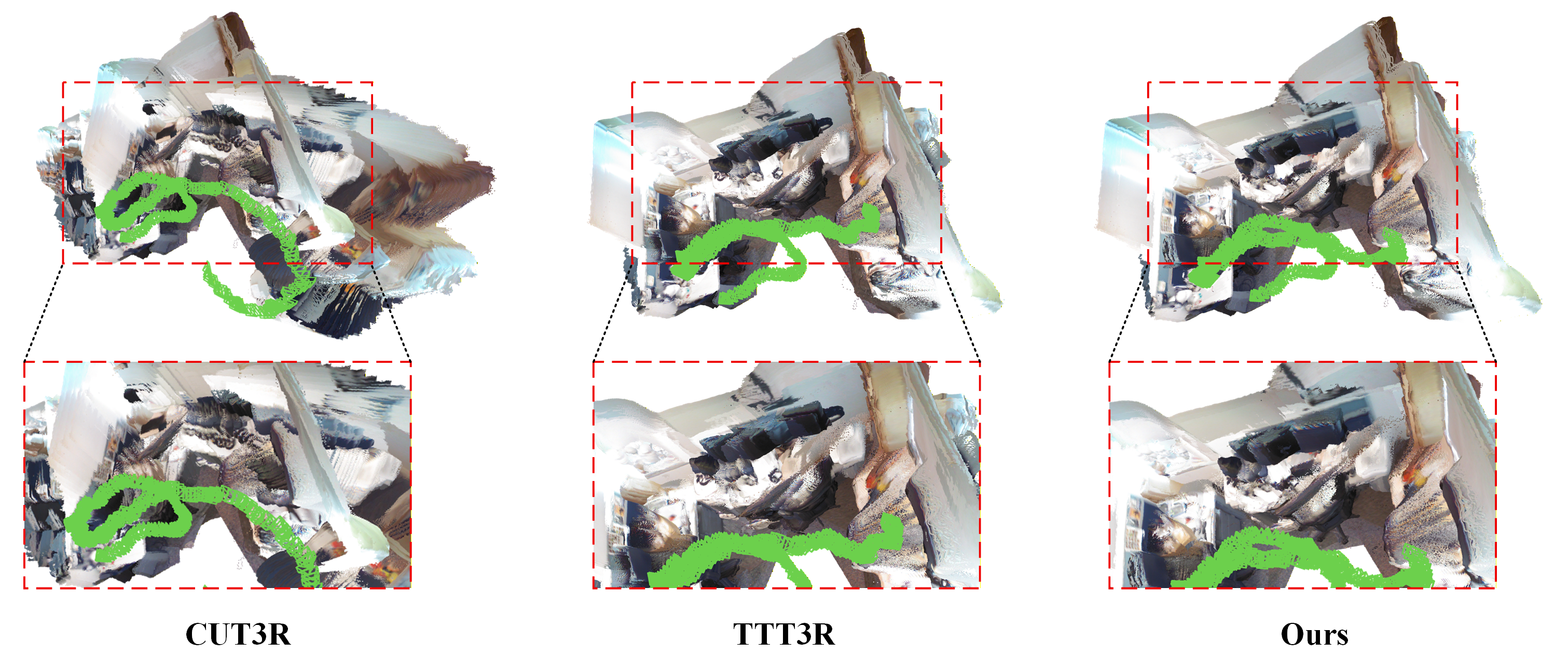}
	\caption{\textbf{Qualitative comparison of 3D reconstruction.} Our method can produce more complete and accurate reconstructions. Please refer to the Supp. Sec.~8 for more visualization comparisons.}
    %\vspace{-0.3cm}
	\label{fig:3DRecon}
\end{figure*}

Figure \ref{fig:3DRecon} further provides qualitative comparisons. CUT3R produces severely distorted geometry and inaccurate camera trajectories. Although TTT3R mitigates these issues, residual artifacts remain visible. Instead, our method generates more coherent reconstructions with accurate scene structure, indicating that selective state updates balance preserving historical information and incorporating new observations.

\begin{figure*}[htbp]
	\centering
	\includegraphics[width=0.65\linewidth]{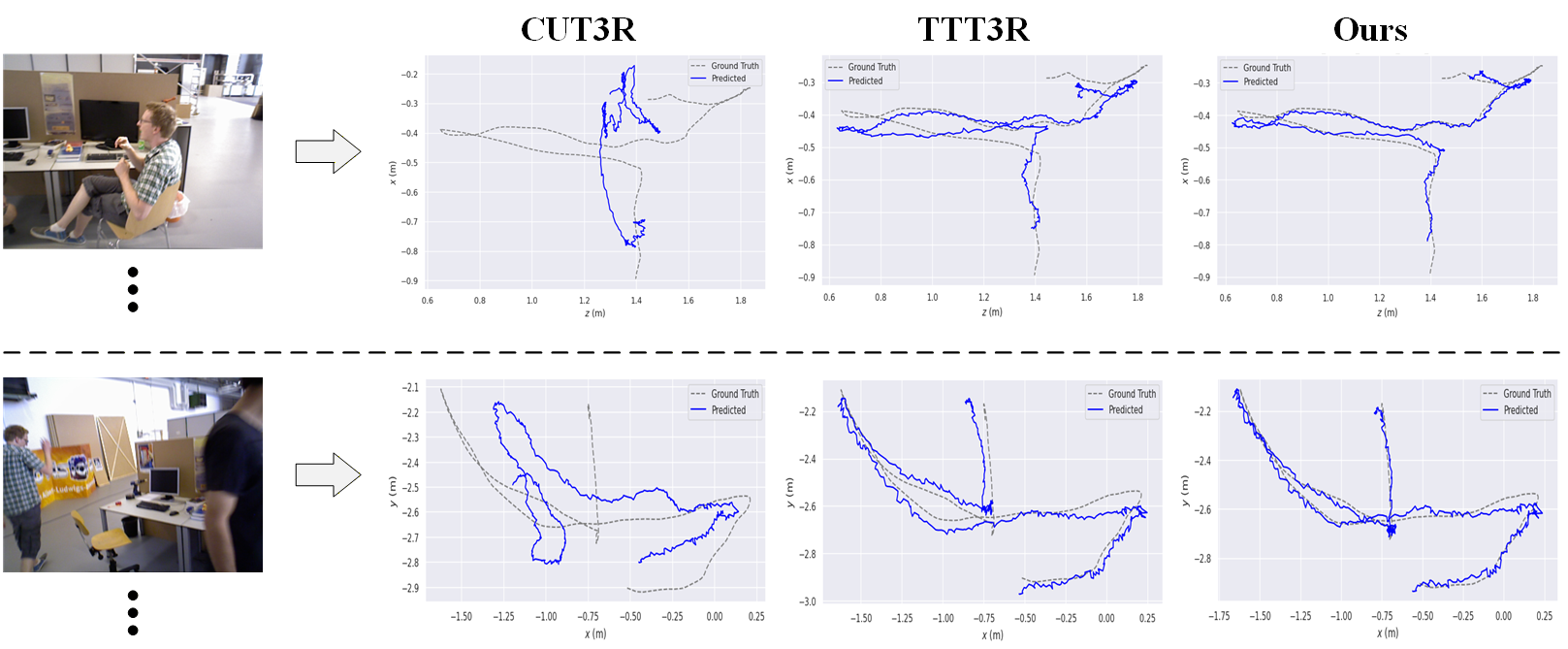}
	\caption{\textbf{Qualitative comparison of predicted camera trajectories} on TUM-dynamics \cite{sturm2012benchmark} dataset. The trajectories are visualized along the two axes with maximum variance. Gray dashed lines show ground truth and blue lines show predictions.}
    \vspace{-0.2cm}
	\label{fig:CTvisual-sitting}
\end{figure*}

\begin{table}[htbp]
\caption{\textbf{Ablation studies of each component.} We evaluate video depth estimation on Bonn \cite{palao2019refun} dataset (per-sequence scale) and camera pose estimation on TUM-dynamics \cite{sturm2012benchmark} dataset.}
\label{table:3}
\centering
\setlength{\tabcolsep}{8pt}
\renewcommand{\arraystretch}{1.2}
\resizebox{0.85\linewidth}{!}{
\begin{tabular}{ccccc|cc}
\toprule[1pt]
\multirow{2}{*}{\textbf{Method}} & \multirow{2}{*}{} & \multirow{2}{*}{} & \multicolumn{2}{c}{\textbf{Bonn}} & \multicolumn{2}{c}{\textbf{TUM-dynamics}} \\ 
\cmidrule(lr){4-5} \cmidrule(lr){6-7}
& & & Abs Rel $\downarrow$ & $\delta<1.25 \uparrow$ & ATE $\downarrow$ & RPE rot $\downarrow$ \\ 
\midrule[0.6pt]
Baseline & & & 0.078 & 93.7 & 0.046 & 0.473 \\
TAUM & \checkmark & & 0.066 & 95.9 & 0.028 & 0.375 \\
SCUM & & \checkmark & 0.074 & 94.1 & 0.040 & 0.415 \\
Ours & \checkmark & \checkmark & \textbf{0.064} & \textbf{96.5} & \textbf{0.026} & \textbf{0.372} \\ 
\bottomrule[1pt]
\end{tabular}
}
\end{table}

\subsection{Ablation Studies and Analysis}

\noindent\textbf{Effect of Each Component.} To assess the contribution of each module, we conduct ablation studies on Bonn \cite{palao2019refun} and TUM-dynamics \cite{sturm2012benchmark} using CUT3R \cite{wang2025continuous} as the baseline. As shown in Table \ref{table:3}, adding TAUM alone reduces Abs Rel from 0.078 to 0.066 and ATE from 0.046 to 0.028, indicating that temporal state evolution effectively mitigates error accumulation, especially for pose estimation. Adding SCUM alone obtains 0.074 Abs Rel and 0.040 ATE by preventing uniform updates across regions. When both modules are integrated, TTSA3R achieves the best performance with 0.064 Abs Rel and 0.026 ATE.

\noindent\textbf{Comparison with State Reset.} Periodic state reset \cite{chen2025ttt3r,cheng2026horizonstream} is a simple strategy to limit state drift. We compare with it on 1000-frame TUM-dynamics sequences using a reset interval of 100 frames (Supp. Sec.~4). It can be seen that reset improves the ATE of CUT3R from 0.174 to 0.124 and that of TTT3R from 0.111 to 0.097. In comparison, TTSA3R achieves 0.092 without any reset and further reaches 0.079 when combined with reset. This shows that our adaptive update reduces trajectory drift more effectively than hard reset in terms of ATE, while remaining complementary to it.

\noindent\textbf{Camera Trajectory Analysis.} We visualize predicted camera trajectories on TUM-dynamics \cite{sturm2012benchmark} dataset in Figure \ref{fig:CTvisual-sitting}, which presents two representative indoor sequences with moving persons. It can be seen that CUT3R \cite{wang2025continuous} accumulates progressive drift and deviates abruptly during rapid viewpoint changes. Though TTT3R \cite{chen2025ttt3r} reduces the drift magnitude, noticeable deviations remain. These observations reflect how existing recurrent methods lack effective adaptive mechanisms that respond to varying observations and temporal dynamics. In contrast, our predicted trajectories closely align with the reference path. Since camera poses are regressed frame by frame, slight local jitter exists in all methods. Nevertheless, TTSA3R achieves the lowest RPE among streaming methods on TUM-dynamics (Table \ref{table:2}) and lower RPE$_{\text{rot}}$ than CUT3R and TTT3R on 1000-frame sequences (0.441 vs.\ 0.548 and 0.452, Supp. Sec.~4), indicating that our adaptive updates reduce drift without introducing additional instability. Additional comparisons on ScanNet are provided in Supp. Sec.~7.

% \begin{figure}[h!]
% 	\centering
% 	\includegraphics[width=0.58\linewidth]{imgs/Comparison.png}
% 	\caption{\textbf{Comparison of inference efficiency.} We compare GPU memory used and FPS of different methods using 512$\times$144 image resolution on KITTI \cite{geiger2013vision} dataset.}
%     %\vspace{-0.3cm}
% 	\label{fig:comparison}
% \end{figure}

% \noindent\textbf{Inference Efficiency Analysis.} We further evaluate computational efficiency by measuring inference speed and GPU memory consumption on video depth estimation. 
% As illustrated in Figure \ref{fig:comparison}, spatial memory-based methods Spann3R \cite{wang20243d} and Point3R \cite{wu2025point3r} suffer from substantial memory overhead. Conversely, CUT3R \cite{wang2025continuous} preserves constant memory but its performance degrades on extended sequences. Similarly, TTT3R \cite{chen2025ttt3r} improves memory efficiency by achieving 6 GB consumption. Notably, TTSA3R achieves the lowest memory footprint at 5 GB with 18.5 FPS inference speed. These results demonstrate that our method achieves a competitive efficiency-memory trade-off among existing streaming methods.
\section{Conclusions}

In this work, we present a training-free 3D reconstruction framework TTSA3R that mitigates catastrophic forgetting of recurrent models for streaming reconstruction through an adaptive temporal-spatial mechanism. Specifically, the key contribution is a dual-module design that models temporal state dynamics and spatial correspondence, enabling complementary selective updates based on cross-frame evolution and observation quality. 
Thus, this approach enables fine-grained control over memory persistence with real-time inference efficiency. Extensive experiments demonstrate that our method achieves the largest gains on extended sequences across depth, pose and reconstruction tasks and remains competitive with state-of-the-art methods on standard short-sequence benchmarks.

\noindent\textbf{Limitations.} Our method is optimized for extended streaming scenarios with adequate visual overlap. Performance degrades under severe occlusions or noisy estimates, where temporal and correspondence signals become unreliable. Additionally, as a training-free design, the adaptive mechanism is limited by the representational capacity of the base recurrent architecture.
\newpage
{
    \small
    \bibliographystyle{ieeenat_fullname}
    \bibliography{main}
}

\clearpage
\twocolumn[
  \begin{center}
    {\LARGE \textbf{Supplementary Material}}
    \vspace{1em}
  \end{center}
]

\setcounter{section}{0}
\setcounter{figure}{0}
\setcounter{table}{0}

\section{Algorithm Details}

In Algorithm \ref{alg:ttsa3r}, we describe the inference procedure of TTSA3R. The framework processes frames sequentially, where the decoder generates candidate states from encoded features and the previous persistent state. For frames beyond the first, TAUM and SCUM compute temporal and spatial masks by evaluating state evolution and spatial interaction respectively. The final update mask is produced by element-wise multiplication of these masks, which selectively updates the persistent state at per-token granularity. This adaptive mechanism helps to preserve stable geometry while integrating new observations across extended sequences.

\begin{algorithm}[htbp]
\caption{Pseudo codes of TTSA3R}
\label{alg:ttsa3r}
\begin{algorithmic}[1]
\State \textbf{INPUT:} Image sequence $\{I_1, \ldots, I_T\}$
\State \textbf{OUTPUT:} Depth maps $\{Z_t\}$, Poses $\{P_t\}$, 3D Pointmaps $\{X_t\}$
\State Initialize state $S_0$
\For{$t = 1$ to $T$}
    \State $F_t \gets \text{Encoder}(I_t)$
    \State $\tilde{S}_t, G_t \gets \text{Decoder}(F_t, S_{t-1})$
    \State $Z_t, P_t, X_t \gets \text{TaskHeads}(G_t, F_t)$
    \If{$t > 1$}
        \State $M_{\text{temp}} \gets \text{TAUM}(\tilde{S}_t, \tilde{S}_{t-1})$
        \State $M_{\text{spat}} \gets \text{SCUM}(F_t, F_{t-1}, \text{Attention}(S_{t-1}, F_t))$
        \State $M_{\text{final}} \gets M_{\text{temp}} \times M_{\text{spat}}$
        \State $S_t \gets M_{\text{final}} \odot \tilde{S}_t + (1 - M_{\text{final}}) \odot S_{t-1}$
    \Else
        \State $S_t \gets \tilde{S}_t$
    \EndIf
\EndFor
\end{algorithmic}
\end{algorithm}

\begin{figure*}[htbp]
	\centering
	\includegraphics[width=1\linewidth]{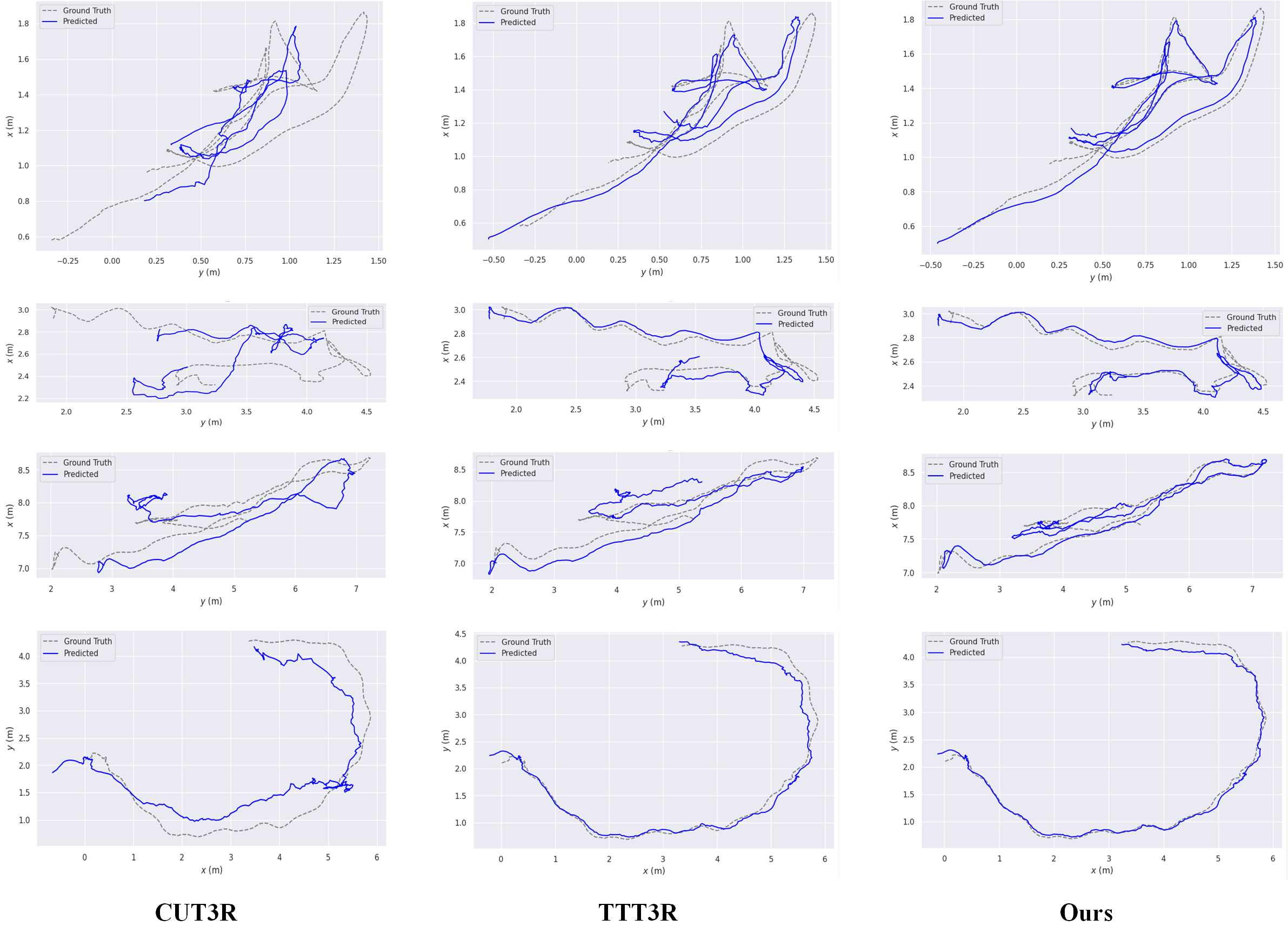}
	\caption{\textbf{Qualitative comparison of predicted camera trajectories} on ScanNet \cite{dai2017scannet} dataset. The trajectories are visualized along the two axes with maximum variance. Gray dashed lines show ground truth and blue lines show predictions.}
	\label{fig:CT-Scannet}
\end{figure*}

\begin{figure*}[htbp]
	\centering
	\includegraphics[width=1\linewidth]{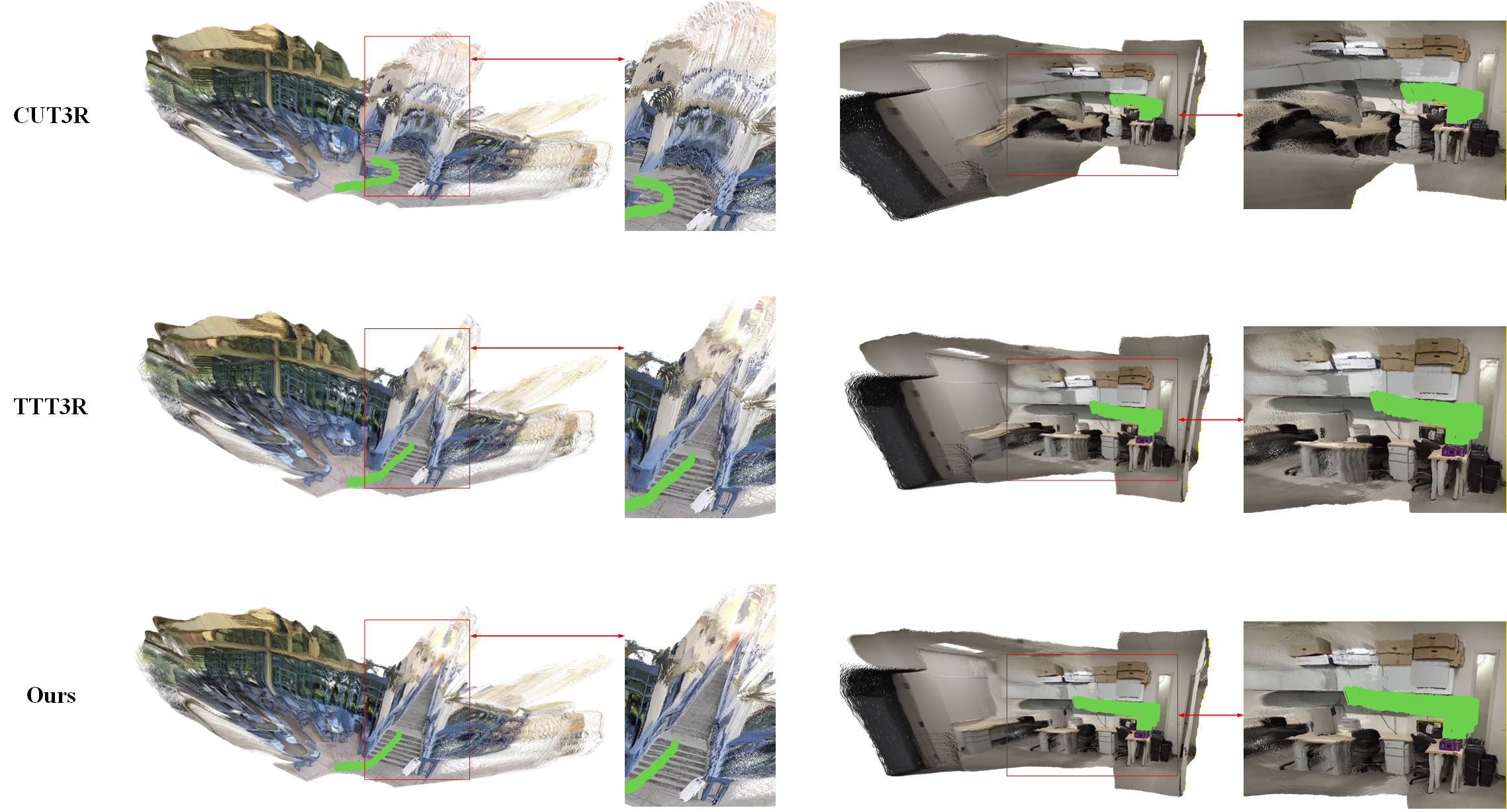}
	\caption{\textbf{Qualitative comparison of In-the-wild Video reconstruction.}}
	\label{fig:Visual-supp}
\end{figure*}

\section{Analysis of mask fusion strategies}

To demonstrate our design choice of mask fusion strategy, we compare four different approaches on the Bonn \cite{palao2019refun} dataset, as shown in Table \ref{table:fusion}. Firstly, we adopt the average strategy that produces intermediate mask values of temporal and spatial masks, and obtain 0.097 Abs Rel and 90.3\% accuracy. Then, we take the maximum of both masks and this results in overly aggressive behavior with 0.102 Abs Rel and 88.7\% accuracy. Furthermore, the minimum strategy achieves 0.085 Abs Rel with 95.7\% accuracy. Since it keeps only the smaller of the two masks for each token, the other signal has no effect on the update. Finally, we adopt element-wise multiplication, where the spatial mask modulates the temporal update weight. This achieves the best performance of 0.079 Abs Rel and 96.6\% accuracy. These results show that multiplication effectively balances preservation of stable geometry and integration of new observations.

\begin{table}[tbp]
\caption{\textbf{Ablation on mask fusion strategies.} We evaluate video depth estimation on Bonn \cite{palao2019refun} dataset using metric scale.}
\label{table:fusion}
\centering
\setlength{\tabcolsep}{8pt}
\renewcommand{\arraystretch}{1.2}
\resizebox{0.95\linewidth}{!}{
\begin{tabular}{c|cccc|cc}
\toprule[1pt]
\multirow{2}{*}{\textbf{Method}} & \multicolumn{4}{c|}{\textbf{Fusion Type}} & \multicolumn{2}{c}{\textbf{BONN}} \\ 
\cmidrule(lr){2-5} \cmidrule(lr){6-7}
& Avg & Max & Min & \textbf{Mul (Ours)} & Abs Rel $\downarrow$ & $\delta<1.25 \uparrow$ \\ 
\midrule[0.6pt]
1 & \checkmark &  &  &  & 0.097 & 90.3 \\
2 &  & \checkmark &  &  & 0.102 & 88.7 \\
3 &  &  & \checkmark &  & 0.085 & 95.7 \\
4 &  &  &  & \checkmark & \textbf{0.079} & \textbf{96.6} \\ 
\bottomrule[1pt]
\end{tabular}
}
\end{table}

\begin{table}[tbp]
\caption{Sensitivity Analysis of  Parameter $\tau$. BONN depth is evaluated under per-sequence scale.}
\label{tab:tau}
\centering\small
\begin{tabular}{ccc|cc}
\toprule
\multirow{2}{*}{$\tau$} & \multicolumn{2}{c|}{\textbf{BONN}} & \multicolumn{2}{c}{\textbf{TUM-dynamics}} \\
 & AbsRel$\downarrow$ & $\delta_{1.25}\uparrow$ & ATE$\downarrow$ & RPErot$\downarrow$ \\
\midrule
0.5 & 0.068 & 95.5 & 0.030 & 0.378 \\
1.0 & 0.064 & 96.6 & 0.027 & 0.373 \\
1.5 & 0.064 & 96.5 & 0.026 & 0.372 \\
2.0 & 0.064 & 96.5 & 0.026 & 0.372 \\
2.5 & 0.064 & 96.5 & 0.026 & 0.378 \\
\bottomrule
\end{tabular}
\end{table}

\section{Sensitivity of $\tau$}

Table \ref{tab:tau} shows stable performance for $\tau\in[1.0,2.5]$, with only $\tau{=}0.5$ showing degradation from over-aggressive updates. $\tau$ is the threshold in Eq.~6 \footnote{Equations are from the main paper} that controls the temporal update gate. The choice $\tau{=}1.5$ is principled: $\hat{\Delta}_t$ is normalized by its global mean (Eq.~5), so $\tau{=}1.5$ assigns larger update weights to tokens whose temporal change is 50\% above the scene average, consistently across all scenes and datasets.

% \begin{table}[tbp]
% \caption{Reset baseline comparison ($N{=}100$).}
% \label{tab:reset}
% \centering\small
% \begin{tabular}{ccc|cc}
% \toprule
% \multirow{2}{*}{Method} & \multicolumn{2}{c|}{\textbf{BONN}} & \multicolumn{2}{c}{\textbf{TUM-dynamics}} \\
%  & AbsRel$\downarrow$ & $\delta_{1.25}\uparrow$ & ATE$\downarrow$ & RPErot$\downarrow$ \\
% \midrule
% CUT3R$+$reset & 0.085 & 94.0 & 0.206 & 2.489 \\
% TTT3R$+$reset & 0.079 & 95.1 & 0.200 & 2.716 \\
% Ours$+$reset  & 0.076 & \textbf{95.5} & 0.200 & 2.729 \\
% \midrule
% CUT3R         & 0.085 & 93.7 & 0.174 & 0.548 \\
% TTT3R         & 0.077 & 95.3 & 0.111 & 0.452 \\
% \textbf{Ours} & \textbf{0.075} & \textbf{95.5} & \textbf{0.092} & \textbf{0.441} \\
% \bottomrule
% \end{tabular}
% \end{table}

% \section{Reset Experiments}

% \cref{tab:reset} compares reset baselines with interval ($N{=}100$) on TUM-dynamics (1000 frames)
% and BONN (500 frames) datasets. The results illustrate that periodic reset introduces severe pose degradation across all methods; for example, CUT3R's RPErot increases from 0.548 to 2.489 after reset. Our method without reset consistently outperforms all these baselines with reset, and Ours$+$reset performs worse than Ours on the TUM-dynamics dataset, indicating that adaptive gating and periodic reset are conflicting strategies. 

\section{Reset Baseline Experiments}

Table \ref{tab:reset} compares our method against reset-based baselines ($R{=}100$) on TUM-dynamics (1000 frames) and BONN (500 frames). We adopt $R{=}100$, the optimal interval identified by TTT3R. We reset the recurrent state to its initial value every $R$ frames and stitch the resulting per-chunk poses into a unified world frame via metric camera poses, without any optimization. Formally, $T_k^{\text{local}}(t)$ denotes the camera pose at frame $t$ within chunk $k$, expressed in that chunk's own reset-origin frame. Let $t_k$ denote the first frame of chunk $k$. We obtain the global pose by chaining consecutive chunks at their boundary:
\begin{equation}
T_k^{\text{global}}(t) = T_{k-1}^{\text{global}}(t_k{-}1) \cdot \big[T_k^{\text{local}}(t_k)\big]^{-1} \cdot T_k^{\text{local}}(t)
\end{equation}
This anchors each new chunk to the last estimated global pose of the previous chunk and exploits the physical continuity of camera motion across the reset boundary. BONN depth is evaluated under metric scale. This evaluation preserves absolute scale and reveals the effect of state drift.

It can be seen that periodic reset consistently improves pose estimation for all baselines. CUT3R's ATE improves from 0.174 to 0.124 and TTT3R's ATE improves from 0.111 to 0.097. This indicates that our implementation closely reproduces the trend reported by TTT3R and that state drift is a genuine problem in long sequences. However, Ours without reset already reaches an ATE of 0.092, lower than 0.124 for CUT3R with reset and 0.097 for TTT3R with reset. This shows that our adaptive gating suppresses state drift more effectively than a periodic hard reset in terms of ATE. Nevertheless, reset further improves our method: BONN metric depth decreases from 0.098 to 0.090 and TUM pose ATE decreases from 0.092 to 0.079. The two strategies are therefore complementary rather than redundant when combined. These results demonstrate that the gains of our method stem from the adaptive update mechanism itself, not merely from limiting state retention over long horizons.

\begin{table}[htbp]
\caption{Reset baseline comparison ($R{=}100$) on long sequences. BONN depth is evaluated under metric scale.}
\label{tab:reset}
\centering\small
\begin{tabular}{ccc|cc}
\toprule
\multirow{2}{*}{Method} & \multicolumn{2}{c|}{\textbf{BONN}} & \multicolumn{2}{c}{\textbf{TUM-dynamics}} \\
 & AbsRel$\downarrow$ & $\delta_{1.25}\uparrow$ & ATE$\downarrow$ & RPErot$\downarrow$ \\
\midrule
CUT3R$+$reset & 0.102 & 90.9 & 0.124 & 0.407 \\
TTT3R$+$reset & 0.094 & 93.4 & 0.097 & \textbf{0.382} \\
\textbf{Ours$+$reset}  & \textbf{0.090} & \textbf{94.5} & \textbf{0.079} & 0.384 \\
\midrule
CUT3R         & 0.100 & 90.6 & 0.174 & 0.548 \\
TTT3R         & 0.101 & 91.9 & 0.111 & 0.452 \\
\textbf{Ours} & \textbf{0.098} & \textbf{93.0} & \textbf{0.092} & \textbf{0.441} \\
\bottomrule
\end{tabular}
\end{table}

\section{Inference Efficiency Analysis}

We evaluate computational efficiency by measuring inference speed and GPU memory consumption on video depth estimation. As shown in Table~\ref{tab:efficiency}, spatial memory-based methods Spann3R~\cite{wang20243d} and Point3R~\cite{wu2025point3r} suffer from substantial memory overhead. Conversely, CUT3R~\cite{wang2025continuous} preserves constant memory but its performance degrades on extended sequences. TTSA3R matches TTT3R's~\cite{chen2025ttt3r} efficiency at 18.0 FPS and 6 GB memory consumption without introducing extra learnable parameters or forward passes. These results demonstrate that our method achieves a competitive efficiency-memory trade-off among existing streaming methods.

\begin{table}[htbp]
\centering
\caption{\textbf{Comparison of inference efficiency.} We compare GPU memory used and FPS of different methods using $512\times144$ image resolution on KITTI~\cite{geiger2013vision} dataset. The best and second best results are marked in bold and underline.}
\label{tab:efficiency}
\begin{tabular}{lcc}
\toprule
Method & FPS $\uparrow$ & GPU Memory (GB) $\downarrow$ \\
\midrule
Spann3R~\cite{wang20243d}  & 13.6 & 11 \\
Point3R~\cite{wu2025point3r} & 5.0  & 10 \\
CUT3R~\cite{wang2025continuous}   & \textbf{20.5} & \underline{8}  \\
TTT3R~\cite{chen2025ttt3r}   & \underline{18.0} & \textbf{6}  \\
Ours          & \underline{18.0} & \textbf{6} \\
\bottomrule
\end{tabular}
\end{table}

\section{Mask Behavior Analysis}
\begin{figure}[tbp]
\centering
\includegraphics[width=\columnwidth]{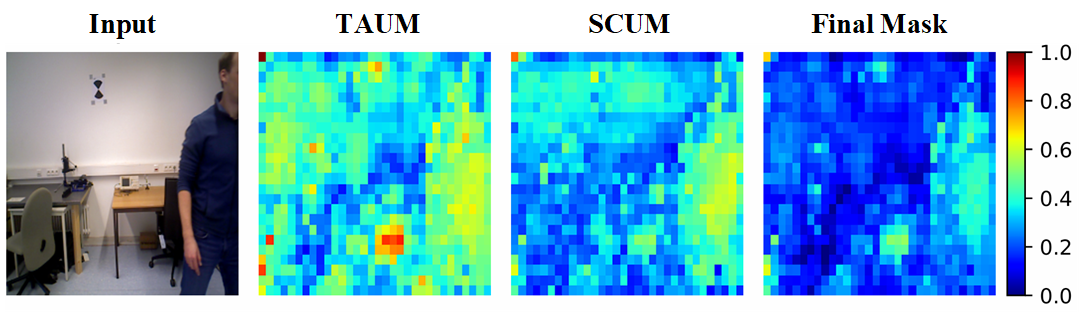}
\caption{Representative Mask Visualization on BONN dataset.}
\label{fig:mask}
\end{figure}

Figure \ref{fig:mask} visualizes TAUM, SCUM, and the final fused mask on a representative frame of a BONN sequence. Each token-level mask is projected to the image plane through its cross-attention to image patches, and all maps share a common min-max normalization, so the colors indicate relative rather than absolute mask values. TAUM produces scattered high-activation tokens at regions with substantial temporal state change. SCUM concentrates activation at attended regions whose observations change across frames. The final mask assigns the largest update weights to tokens where both temporal and spatial signals are high, which shows a sparser and more targeted update pattern than either individual mask. This matches the intended design of multiplicative fusion, where the spatial mask attenuates temporal updates that lack spatial relevance.

\section{More Camera Trajectory Comparisons}

To further illustrate the robustness of our method on extended sequences, we present additional camera trajectory comparisons on ScanNet \cite{dai2017scannet} dataset in Figure \ref{fig:CT-Scannet}. The results reveal distinct performance patterns: uniform updates lead to progressive error accumulation, while attention-based adaptation improves stability yet remains insufficient for complex motion scenarios. Specifically, severe drift becomes evident in CUT3R \cite{wang2025continuous} trajectories as sequences extend, with deviations particularly pronounced in the first, second and fourth scenes featuring intricate camera movements. TTT3R \cite{chen2025ttt3r} reduces this drift but cannot eliminate deviations in challenging scenarios. Our framework TTSA3R with temporal-spatial adaptive updates successfully maintains trajectory accuracy throughout these sequences, demonstrating its effectiveness in alleviating long-term catastrophic forgetting.

\section{More 3D Reconstruction Visualizations}

More 3D reconstruction visualizations on in-the-wild videos are provided in Figure \ref{fig:Visual-supp}. The results reveal distinct failure patterns of the baseline methods. In the left column showing a courtyard scene, CUT3R \cite{wang2025continuous} exhibits severe geometric collapse with fragmented staircase surfaces and TTT3R \cite{chen2025ttt3r} partially recovers the structure but warped steps remain visible. The right column presents an indoor office scene where CUT3R similarly corrupts geometry and merges furniture into irregular boundaries. TTT3R improves surface quality but cannot fully eliminate edge inconsistencies. Our method retains structural integrity across both scenarios and accurately preserves fine-grained details throughout the sequences.

\section{Long-Sequence Evaluation Protocol}

We follow the long-sequence protocol of TTT3R \cite{chen2025ttt3r}. For each target length, we take the corresponding number of leading frames of every test sequence, so that results at different lengths are evaluated on prefixes of the same sequences. Sequences shorter than the target length are used in full.

For video depth estimation (Figure 4 of the main paper), we use the five Bonn \cite{palao2019refun} sequences (balloon2, crowd2, crowd3, person\_tracking2 and synchronous), starting from frame 30 with a stride of 1 and lengths ranging from 50 to 500 frames. Depth is evaluated under metric scale, and the metrics are averaged over sequences weighted by the number of valid pixels. For camera pose estimation (Figure 5 of the main paper), we use all TUM-dynamics \cite{sturm2012benchmark} sequences with a stride of 1 and the ScanNet \cite{dai2017scannet} test scenes with a stride of 3, where the lengths range from 50 to 800 frames. ATE and RPE are computed after Sim(3) alignment and averaged over sequences. For 3D reconstruction (Figure 6 of the main paper), we sample every second frame of each NRGBD \cite{azinovic2022neural} scene with lengths ranging from 50 to 250 frames, and average Acc and NC over scenes. The reset experiments in Sec.~4 follow the same protocol with 1000-frame TUM-dynamics and 500-frame Bonn sequences.

\begin{table}[t]
\caption{\textbf{Comparison of gating signals} on CUT3R. BONN depth is evaluated under per-sequence scale.}
\label{tab:gate}
\centering\small
\resizebox{\columnwidth}{!}{
\begin{tabular}{l|cc|cc}
\toprule
\multirow{2}{*}{Gating signal} & \multicolumn{2}{c|}{\textbf{BONN}} & \multicolumn{2}{c}{\textbf{TUM-dynamics}} \\
 & AbsRel$\downarrow$ & $\delta_{1.25}\uparrow$ & ATE$\downarrow$ & RPErot$\downarrow$ \\
\midrule
None (CUT3R) & 0.078 & 93.7 & 0.046 & 0.473 \\
Alignment confidence (TTT3R) & 0.069 & 95.4 & 0.028 & 0.380 \\
Temporal evolution (TAUM) & 0.066 & 95.9 & 0.028 & 0.375 \\
TAUM + SCUM (Ours) & \textbf{0.064} & \textbf{96.5} & \textbf{0.026} & \textbf{0.372} \\
\bottomrule
\end{tabular}
}
\end{table}

\section{Comparison of Gating Signals}

To isolate the effect of the temporal signal, we compare different gating signals within the same framework in Table \ref{tab:gate}. TTT3R \cite{chen2025ttt3r} and TAUM are both training-free modifications of the same pretrained CUT3R \cite{wang2025continuous} model, and they differ only in the signal used to compute the per-token update weights. TTT3R derives the weights from the alignment confidence between the state and the observation at the current timestep, whereas TAUM derives them from the candidate-state evolution across consecutive frames. It can be seen that replacing the gate of TTT3R with TAUM reduces Abs Rel from 0.069 to 0.066 and RPE$_{\text{rot}}$ from 0.380 to 0.375, while achieving the same ATE of 0.028. This indicates that the cross-frame temporal signal provides a more informative update criterion than the single-timestep alignment signal. Combining TAUM with SCUM further improves all metrics, reaching 0.064 Abs Rel and 0.026 ATE. The spatial distributions of the temporal, spatial and final masks are visualized in Sec.~6.

\end{document}